\documentclass[sigconf]{acmart}

\acmDOI{}
\acmISBN{}
\acmArticle{}
\acmPrice{}

\acmConference[recsysXfashion'19]
{Workshop on Recommender Systems in Fashion, 13th ACM Conference on Recommender Systems}
{September 20, 2019}
{Copenhagen, Denmark}
\acmYear{2019}
\copyrightyear{2019}

\usepackage{tabularx}
\usepackage{fancyvrb}
\usepackage{pbox}

\newcommand\Tstrut{\rule{0pt}{4ex}}       
\newcommand\Bstrut{\rule[-3ex]{0pt}{0pt}} 
\newcommand{\TBstrut}{\Tstrut\Bstrut} 

\AtBeginDocument{%
  \providecommand\BibTeX{{%
    \normalfont B\kern-0.5em{\scshape i\kern-0.25em b}\kern-0.8em\TeX}}}

\begin{document}

\title{Analyzing Customer Feedback for Product Fit Prediction}

\author{Stephan Baier}
\email{s.baier@reply.de}
\affiliation{%
  \institution{Data Reply GmbH}
  \city{Munich}
  \state{Germany}
}


\begin{abstract}
One of the biggest hurdles for customers when purchasing fashion online, is the difficulty of finding products with the right fit. In order to provide a better online shopping experience, platforms need to find ways to recommend the right product sizes and the best fitting products to their customers. These recommendation systems, however, require customer feedback in order to estimate the most suitable sizing options. Such feedback is rare and often only available as natural text. In this paper, we examine the extraction of product fit feedback from customer reviews using natural language processing techniques. In particular, we compare traditional methods with more recent transfer learning techniques for text classification, and analyze their results. Our evaluation shows, that the transfer learning approach ULMFit is not only comparatively fast to train, but also achieves highest accuracy on this task. The integration of the extracted information with actual size recommendation systems is left for future work.

\end{abstract}

\maketitle

\section{Introduction}

Online fashion retail has gained enormous popularity over the last years. It provides a convenient alternative to traditional brick and mortar fashion shops. However, many customers still shy away from purchasing fashion online due to the difficulty of choosing properly fitting products. Indeed, it has been shown that for a majority of customers, product fit is the most prevalent factor for a satisfying online fashion shopping experience \cite{fit_study}. 

The difficulty of finding products with the right fit, leads to an enormous amount of returns, which are both, expensive for the retailer and harmful to the environment. Recent research has been concerned with the recommendation of products based on the best fit \cite{size_recommendation_1, size_recommendation_2, size_recommendation_3, dataset}. However, these systems need the feedback of customers in a structured form, to estimate both, the customer's preference of fit and an item's true size. This information is often unknown. However, sometimes customers give hints in their review when returning an article. For the online retailer, this is a valuable source of information which is worth to explore, as it enables the downstream recommendation task.

Although, some retailers established the possibility to give feedback about the fit in a review form, customers prefer to phrase their feedback into natural text. Therefore, it is necessary to extract the information from natural text. In this paper we examine the extraction of this hidden information from customer review texts, using natural language processing (NLP) techniques. A typical analysis task on customer reviews is sentiment analysis, which is usually framed as a text classification problem. The problem of fit prediction is  very similar to sentiment analysis and can also be formulated as a supervised text classification problem. Given an input text the model needs to predict whether the ordered garment did fit, or whether it was too large, or too small.

In our experiments we compare different approaches to text classification, which are also commonly used for sentiment analysis. We thereby compare more traditional approaches, such as a linear classifier on top of TF-IDF features with very recent transfer learning methods, namely ULMFit \cite{ulmfit} and BERT \cite{bert}. Transfer learning is a prevalent trend in current NLP research. The idea is that the models are already pre-trained on a large corpus of text, and thus capture a large deal of linguistic information. The models are then fine-tuned in a few epochs of training to the task at hand. We conduct experiments on two real-world datasets  which where recently published in \cite{dataset}.

The paper is structured as follows. In the next section, we discuss related work. Section \ref{section:background} gives an overview of the text classification methods used in this paper. Section \ref{section:experiments} contains a description of the experiments and an analysis of the results. We conclude our work in Section \ref{section:conclusion}.

\section{Related Work}

Product fit recommendation has only been researched very recently. The main challenge is to estimate the true size of a product and the best fitting size for a customer, and match them accordingly. This has been handled in a number of different ways. In \cite{size_recommendation_1} the true size for customers and products is estimated using a latent factor model, and recommendations are made on a similarity-based approach. In \cite{size_recommendation_3}  an extension using a Bayesian model has been proposed. A hierarchical Bayesian approach can be found in \cite{size_recommendation_4}. In \cite{dataset} the size recommendation problem is tackled by learning embeddings for customers and products. The embeddings are combined in a joint space, where metric learning and prototyping is applied in order to derive good representations for the different size classes. The authors of \cite{dataset} also published two datasets with their paper, which we utilize in our experiments. 

In this work, we do not focus on the matching of products and customers itself, but on the extraction of size information from natural text. This is an upstream task, in which the size feedback is translated from unstructured text to structured labels. The problem can also be related to sentiment analysis, which is the most popular approach when analyzing customer reviews. The standard sentiment analysis approach is formulated as a classification problem with three classes (positive, negative, and neutral). A more advanced extension of this is aspect-based sentiment analysis, where sentiments are extracted for various aspects (e.g. various categories, or various entities which are mentioned in the text). In this way, fit prediction can be understood as a particular aspect of sentiment analysis. A recent review on sentiment analysis can be found in \cite{sentiment_analysis_survey}. 

Over the last decades a vast number of techniques have been proposed  for text classification and sentiment analysis. Traditional approaches often use bag-of-word representations in combination with standard classifiers. More recently, learned embeddings, e.g. \cite{glove, flair} have become a popular alternative to bag-of-word representations. Deep learning has become a standard vehicle for a wide variety of NLP tasks. Also a number of transfer learning models have shown state-of-the-art results on text classification \cite{ulmfit, bert, xlnet}. A more detailed discussion on the techniques used in this paper follows in Section \ref{section:background}.

\section{Background}
\label{section:background}

Text classification is a standard machine learning task within natural language processing. Over the last decades, the research community has come up with a large number of approaches for tackling this problem. In this section, we introduce the methods that are applied in the experiments for this paper.

\subsection{Bag-of-Words}

In the bag-of-word model, a text (sentence, paragraph, or whole document) is represented by a fixed length vector, where each dimension of the vector represents a word in the vocabulary. The entries of the vector are usually counts of how often a specific word appears in the text. However, simple counts are not necessary representative of the importance of a word. Thus, term frequency - inverse document frequency (TF-IDF) is a better representation, where the counts are divided by the frequency of the word in the whole corpus. In this way, very common words across the corpus receive less importance. In the bag-of-words representation, the word order, grammar, and other linguistic features are ignored. When using bag-of-words for text classification, the representation vector is used as input to a standard machine learning model, e.g. Linear Regression, Naive Bayes. 

\subsection{Embeddings}

Word embeddings are fixed length vector representations for words, where the similarity between two vectors relates to semantic similarity between the corresponding words. Word embeddings can be derived in many ways, however in recent years, the training of word embeddings as part of the input layer of a neural network has become popular. Therefore, a task is defined for which lots of training data is available. The embeddings are then a by-product of the model for that task. One of the methods which popularized word embeddings, is Word2Vec \cite{word2vec}; Word2Vec is trained on the task of predicting a word, given its surrounding context. Another popular method for word embeddings is GloVe (Global Vectors) \cite{glove}. GloVe vectors are trained on global word co-occurrence statistics of a corpus. The word vectors are solely trained on a large text corpus, e.g. wikipedia, and no additional labels are required. 

When using word embeddings for the representation of text passages, they need to be combined into a fixed length vector. A standard technique is to compute the element-wise mean between the embeddings in order to represent a piece of text, such as a sentence or paragraph.  However, in this way the order of the words is ignored in a similar way as in the bag-of-words approach. A more advanced option is to feed the embeddings into a Recurrent Neural Network (RNN), which processes the sequence and provides hidden representations at each step. Either the last hidden representation of the RNN (usually using an extension such as LSTM or GRU) is then used as a representation of the sequence, or again the mean is taken from the hidden representations at each position of the sequence. The representations of the RNN should then, however, reflect the order of the words in the text. The parameters of the RNN need to be trained on a specific task, e.g. classification.

\subsection{ULMFit}

ULMFit \cite{ulmfit} is a popular transfer learning approach, mainly used for text classification. Transfer learning has become a standard technique in computer vision and is now more and more finding its path also into NLP research. The idea in transfer learning is that large models are pre-trained on large amounts of data. In NLP these models can usually be trained in an unsupervised way, using auxiliary tasks such as predicting the next word. The pre-trained models are then fine-tuned on a downstream NLP task, e.g. text classification. This can be achieved with a much smaller number of training epochs (training a model from scratch, often needs days to converge) and with a relatively small amount of labels. ULMFit trains a AWD-LSTM \cite{awd-lstm} language model which is a specific type of RNN architecture. The model learns to predict the next word given a sequence of previous words. In this way, the model captures many nuances of natural text. The text representation produced by the language model can then be used as a representation for a downstream task, e.g. text classification. The authors of ULMFit propose not only the model architecture but also a detailed procedure on how to fine tune it on a custom task. In this procedure the language model is first fine-tuned on the new data until it converges. Then the classification model is trained given the representation of the language model. For this step, the authors propose the following techniques: The first technique is discriminative fine tuning, which means that when updating the weights based on the new data, different learning rates are used for the different layers $\eta^{l-1} = \eta^l / 2.6$. In this way, the layers closer tho the output are updated by a larger ratio, than the layers closer to the input. The second technique are slanted triangular learning rates, a learning rate schedule which increases fast in a short period and decreases slowly over a long period while training on the new data. The third technique is gradual unfreezing, which means that not all layers are trained at once. First, only the last layer is updated and all other layers are kept constant. Then the second last layer is also added to the trainable parameters, and so on. Finally, the whole model is updated. This strategy prevents catastrophic forgetting of the already learned aspects in the pre-trained language model.

%

\subsection{BERT}

As opposed to ULMFit which uses an AWD-LSTM for the language model, BERT (Bidirectional Encoder Representations from Transformers) \cite{bert} builds on top of transformers, which are attention-based neural network building blocks, recently proposed in \cite{transformer}. One of the main advantages of transformers, besides their excellent modeling power, is that they are easily parallelizable as opposed to RNNs.

The BERT model is pre-trained on a large text corpus, on two tasks: predicting randomly masked words in the input, and predicting for a pair of sentences, if the second sentence is the subsequent sentence to the first one. The pre-trained model can be applied to input texts in order to derive contextualized word embeddings or to derive a representation of the whole sequence. At the output, BERT provides a vector representation for each position in the input. Each input sequence starts with a special token [CLS]. The representation at the position of this token is used as a representation for the whole sequence. When running text classification on top of BERT, this representation is passed to a linear classifier. During the fine tuning all layers of BERT are updated given the data of the downstream task. Google released a number of different pre-trained BERT models. The smallest contains 110 million parameters and the largest 340 million parameters.

\section{Experiments}
\label{section:experiments}

In this section, we describe the datasets and procedures which we use, to evaluate different NLP techniques for the problem of product size prediction from natural text.
\subsection{Datasets}

\begin{table}
	\centering
	\resizebox{\columnwidth}{!}{%
		\begin{tabular}{|l || c | c | c |} 
			\hline
			& Datapoints  & Avg. Tokens & Vocab size \\ [0.5ex] 
			\hline\hline
			ModCloth dataset  & 76059 & 38  & 22110 \\ 
			RTR dataset & 192523 & 53 & 28306 \\
			\hline
		\end{tabular}
	}
	\caption{Statistics of the two datasets that are used in the experiments.}
	\label{table:dataset_stats}
\end{table}

The authors of \cite{dataset} published two datasets for product size recommendation. The first dataset is gathered from the fashion platform \emph{ModCloth}, the second from \emph{RentTheRunway} (RTR). Both datasets contain customer feedback on the product size of fashion items. The feedback is available as structured information, and in form of a feedback statement in natural text. Thus, these datasets are the perfect source for us to train our classification models. Table \ref{table:dataset_stats} summarizes the key statistics of the two datasets. Both datasets consist of three different labels.  In the ModCloth dataset, there are 52,222 datapoints labeled as "fit", 11,717 labeled as "small", and 12,120 labeled as "large". In the RTR dataset, there are 142,042 datapoints labeled as "fit", 25,776 labeled as "small", and 24,705 labeled as "large". 

In our experiments the data can be used to formulate a classification task, where input is the review text and output is one of three classes: "fit", "large", or "small". For our experiments, we split the data into 80 \% training and 20 \% test data. 5 \% of the training data are used as a validation set, for determining the optimal hyper-parameters for each method. We also apply early stopping, which means, that we stop model training once the validation loss does not decrease on the validation set.

\subsection{Results}

\begin{table}
	\centering
	\resizebox{\columnwidth}{!}{%
		\begin{tabular}{|l || c | c |} 
			\hline
			Method & ModCloth dataset & RTR dataset \\ [0.5ex] 
			\hline\hline
			Majority Class & 0.6892 & 0.7396 \\ 
			TF-IDF LR & 0.7899& 0.8033  \\ 
			Mean GloVe LR & 0.7124 & 0.7471 \\
			ULMFit Fine-Tuned & \bf{0.8269} & \bf{0.8420}  \\ 
			BERT Fine-Tuned & 0.8113 & - *  \\ 
			\hline
		\end{tabular}
	}
	\caption{Micro-F1 scores on the test set for the  text classification problem on two different datasets. * We stopped fine tuning BERT on the RTR dataset after the first epoch, which already took 5 hours on a NVIDIA Tesla K80. }
	\label{table:classification_results}
\end{table}


We train a number of models on the classification task.  Table \ref{table:classification_results} shows the results of the different classification models on the test set for both datasets. We report the micro-F1 scores (same as accuracy in multiclass classification). To put the model results into relation, we report the accuracy of the dummy classifier, which always predicts the majority class ("fit") independent of the input. Although, there is some imbalance in both datasets, it was not necessary to apply up- or down-sampling techniques when training the classifiers.

The first model is a linear classifier on top of TF-IDF representations. The dimensionality of the representation vectors is in both datasets the size of the vocabulary, as shown in Table \ref{table:dataset_stats}. We tried to reduce the dimensionality of the vectors by wiping out most frequent words, but could not see any improvement on the classification score. We further train a model, which uses the mean of pre-trained GloVe embeddings of the contained words as a representation for a customer review. This representation is again passed to a linear classifier. The dimensionality of the embeddings is 100. Surprisingly, the results using the mean GloVe vectors are much worse than the TF-IDF features. The TF-IDF representation is very sparse, thus if a keyword is represented in the input, it can easily be picked up by the classifier, whereas in the mean embedding approach the averaging might dilute this information.


We train two transfer learning models, namely ULMFit and BERT. ULMFit achieves the best results on both datasets with 0.8269 and 0.8420 respectively. We tried both bidirectional and uni-directional LSTMs for the language model, however found that the bidirectional model did not improve the results. For both datasets, we found an initial learning rate of 0.02 and a batch size of 32 to work best. We trained the ULMFit language model for 20 epochs on both datasets. We stopped training of the ULMFit classifier after 8 epochs on the ModCloth dataset and after 9 epochs on the RTR dataset. We use the uncased BERT base model which contains 110 million parameters, which is the smallest among the models released by the authors. ULMFit on the other hand only has 36 million parameters. We found a learning rate of 0.00006 for BERT to perform best on the ModCloth dataset, and used a batch size of 5, as we would otherwise run out of memory on the GPU. All other parameters where left at default. The training time for one epoch of ULMFit is about 15 minutes on the RTR dataset, whereas one epoch of BERT already took 5 hours on a NVIDIA Tesla K80 GPU. We stopped the training here as this long runtime did not allow for proper hyperparameter tuning.  On the ModCloth dataset, we trained BERT for 8 epochs.

%
%
%
%
%

The language model of ULMFit can also be used to generate text, as it is trained on the task of predicting the next word given a sequence of past words. The next word is randomly sampled from the distribution predicted by the model. The generated text should be similar to the data it was trained or fine-tuned on. Table \ref{table:text_generation} shows random examples of generated text after giving the model a few seed words for the start of the sentence (shown in bold). The sentences starting with words in green are produced by the fine-tuned model on the ModCloth dataset. The sentences starting with words in red are produced by the language model pre-trained on wikitext. It can be seen very clearly, that the language model has adopted to the new data, as it produces almost realistic review texts.

\begin{table}
	
	\definecolor{custom_green}{rgb}{0.0, 0.5, 0.0}
	\definecolor{custom_red}{rgb}{0.8, 0.25, 0.33}
	
	\resizebox{\columnwidth}{!}{%
		\begin{tabular}{|l|} 
			\hline
			
			\pbox{\columnwidth}{\textbf{\color{custom_green} I have been} looking at this. i've never been so disappointed i didn't even feel like i'll ever keep this cardigan from my wardrobe.} \TBstrut\\
			\hline 
			\pbox{\columnwidth}{\textbf{\color{custom_red} I have been} there for five and a half years ( OK , 18 ] and TV years since ) , has been a hard rock show. } \TBstrut\\
			\hline 
			\pbox{\columnwidth}{\textbf{\color{custom_green} These are} my favorite leggings met your current size, and prego years have held up to multiple washes.}\TBstrut\\
			\hline 
			\pbox{\columnwidth}{\textbf{\color{custom_red} These are} two of the first UK mix settings over the age of one edition . They were given to the older boys at the British Teen.}\TBstrut\\
			\hline 
			\pbox{\columnwidth}{\textbf{\color{custom_green} I own this}  dress in every color. It fits perfectly! Fits super comfortable.}\TBstrut\\
			\hline 
			\pbox{\columnwidth}{\textbf{\color{custom_red} I own this} party = is a British country record . It is called the King's Favourite King with King George II.}\TBstrut\\
			\hline 
			\pbox{\columnwidth}{\textbf{\color{custom_green} This is a} great top. Very flattering with a lot of big waists.}\TBstrut\\
			\hline 
			\pbox{\columnwidth}{\textbf{\color{custom_red} This is a} legitimate way of finding things in the public mind too.}\TBstrut\\
			\hline 
		\end{tabular}
	}
	\caption{Random examples of generated text, using the AWD-LSTM language model trained with ULMFit. The bold words where given to the model as a sentence start, and the remaining words where generated by the model. The green sentences are generated by the fine-tuned language model, whereas the red sentences where generated by the pre-trained model.}
	\label{table:text_generation}
\end{table}

\section{Conclusion}
\label{section:conclusion}

Extracting product fit information from customer reviews is an important problem in online fashion retail. We applied various text classification models to the task of predicting product fit, given customer feedback. The experiments show that a standard TF-IDF approach achieves already quite good results, but also that recent transfer learning approaches can boost the performance significantly. ULMFit was found to achieve best performance. It is also relatively fast to train, compared to BERT, which in the small version has about three times more parameters than ULMFit and in the large version has almost 10 times more parameters. We conclude that most of the large pre-trained language models, which have gained a lot of attention recently (BERT is just one example, others are FLAIR \cite{flair}, ELMo \cite{elmo}, or XLNet \cite{xlnet}) might be too over-sized for the comparatively simple problem of product fit classification, which we examined in this paper.

Next steps include the application of the trained models to more general review texts, where the customer was not specifically asked for product fit. Here, it might be beneficial to introduce a fourth class, which represents the case, that no product size information is included. Also more fine-grained models, predicting which part of the garment did not fit, would be desirable. Another important aspect to consider is, that for a large amount of fashion products no review data exists.  Similarly, most users do not give explicit feedback about the fit of their purchased and returned items. Thus, for future work it is desirable to train a model which relates item or user features to product sizes and preferences. Such a model can either be based on other available user and item features, or based on collaborative filtering, which exploits the interactions between users and items given the implicit feedback of purchased and returned items. The extracted or inferred information about product fit can then be used in manifold downstream tasks, such as improving search result rankings, pre-selecting the recommended product size in the purchase order, or simply informing customers about the fit of a specific product.

%
%
%
%
%
%

\bibliographystyle{ACM-Reference-Format}
\bibliography{references}

\end{document}